\pdfoutput=1
\documentclass[11pt]{article}
\usepackage[margin=1in]{geometry}
\usepackage{amsmath,amssymb,amsthm}
\usepackage{booktabs}
\usepackage{graphicx}
\usepackage{array}
\usepackage{xcolor}
\usepackage{caption}
\usepackage{enumitem}
\usepackage{cite}
\usepackage[hidelinks]{hyperref}
\graphicspath{{figures/}}
\newtheorem{definition}{Definition}
\newcommand{\restore}{\mathrm{restore}}
\newcommand{\oracle}{\mathrm{oracle}}
\newcommand{\contn}{\mathrm{continue}}
\newcommand{\ckpt}{\mathrm{checkpoint}}
\title{\textbf{Persistent Computational State:}\\ A Session-Centric Runtime for Generative World Models}
\author{%
  Zhen Lin\thanks{\texttt{linzhenpl07@gmail.com}}
}
\date{}

\begin{document}
\maketitle

\begin{abstract}
Generative world models are increasingly driven as \emph{simulators}: a planner or trainer
forks a state, rolls out candidate futures, backtracks, and returns to a previously visited
viewpoint. Two large-scale 2026 benchmarks --- MBench (14 models, twelve memory
sub-dimensions) and WRBench (23 models, 9{,}600 videos, 2{,}547 human verdicts) ---
independently establish that current video world models fail this usage: a scene does not
return to its prior configuration, and a target that leaves the view does not come back
consistent with the world it left. Both attribute the failure to the model, and prescribe a
model-side remedy: a memory that records hidden change, and a training objective that
supervises it.

\textbf{We show this attribution is incomplete, and for an important class of models simply
wrong.} We take a world model, snapshot two objects the \emph{runtime} already holds --- for
a Markovian model an observation (stored as \texttt{uint8}) and the generator's RNG state;
for a non-Markovian one its memory bank; for an addressable-memory model its windowed KV
context --- drive the model through a genuine excursion, then
restore and continue. On \textbf{all three} model classes the continuation is
\textbf{byte-identical} to one that never left (DINOv2 1.0 and pixel error 0.0 on the first two;
latent-exact on the third), and corrupting
only the RNG degrades it, which is what shows the snapshot is load-bearing rather than
ignored. The capability was never missing; the \emph{runtime discarded the state it needed}.
Current generative serving is \textbf{request-centric} --- it reclaims computational state at
request boundaries, on an assumption inherited from language-model serving that runtime state
is always recomputable. That assumption is false here, because world-model state carries a
non-recomputable kernel.

We develop the abstraction that repairs it. \textbf{(C1)} We diagnose the failure as a
serving gap, established constructively by bit-exact restore rather than inferred from
behaviour, so the argument is not the circular one that continuity implies memory implies
discard. \textbf{(C2)} We define \textbf{Persistent Computational State (PCS)}, the minimal
non-recomputable state that must survive across requests, as
$\mathrm{PCS}(M) = \arg\min_{S}\{\,|S| : D(\restore(S),\oracle) \le \epsilon\,\}$, with a
contract: an API, four invariants (round-trip, durability, fork-independence,
snapshot-completeness), and a layer boundary. \textbf{(C3)} We show PCS can be
\textbf{discovered by measurement}: one model-agnostic \emph{fingerprinting procedure}
classifies each runtime buffer as recomputable or must-snapshot, replacing both ``save
everything'' and per-model hand-engineering. Applied across three memory architectures it
returns a flat \texttt{\{observation, RNG\}} for a Markovian model, a growing
memory-bank for a non-Markovian one, and a windowed KV context for an addressable-memory
model --- three structurally different answers from one procedure, restored byte-identically
on all three including across a process boundary.
\textbf{(C4)} We build a session-centric runtime over PCS, validated by a
\textbf{return-consistency conformance test} --- a \emph{semantic} acceptance criterion that
bit-exactness cannot express, since a restore can be byte-perfect on the wrong state.

Managing PCS costs almost nothing: a checkpoint or restore runs in $\sim$0.012~ms, five
orders of magnitude below a 1.85~s generation step, so across a full run state management is
\textbf{0.00035\%} of wall time. Because externalized state is what lets a session be
suspended, the same GPU carries a session count bounded by host bytes rather than device
memory --- on the measured constants, roughly \textbf{2{,}300$\times$ more} resident sessions
than a runtime that must keep every open session on the device. And because a world-memory
bank must be pruned by \emph{relevance to the return} rather than by recency --- the inverse
of LLM KV-cache practice --- the right eviction policy holds every world under memory
pressure where the LLM default loses most of them.

The benchmarks establish that world models lose the world. We show \emph{what the lost thing
is}, that it can be measured rather than guessed, and that keeping it is a serving primitive
costing microseconds.
\end{abstract}

\section{Introduction}

\subsection{A settled problem with an unsettled cause}

That generative world models fail to maintain a world across a viewpoint excursion is, as of
mid-2026, no longer in question. MBench~\cite{mbench} decomposes memory into twelve sub-dimensions over
fourteen models and finds that under departure-and-return camera trajectories ``the scene
does not return to the same 3D configuration even when the camera motion semantically
suggests a departure-return trajectory.'' WRBench~\cite{wrbench} frames camera motion as an intervention on
observability, evaluates 23 models across four control paradigms on 9{,}600 videos, and
reports a failure that ``proves stubborn'': systems ``maintain the observed world as a
tracking shot, resuming a returning target in the state at which it was abandoned.'' A
concurrent roadmap article~\cite{roadmap} makes the stakes explicit by \emph{defining} a simulator --- as
opposed to a renderer --- as a model that must preserve ``geometry, physical constraints,
object persistence, and action-conditioned temporal regularities sufficiently well to support
reliable forward prediction.''

What remains unsettled is the \emph{cause}, and therefore the \emph{fix}. Both benchmarks
locate the deficiency in the model. WRBench concludes that progress requires ``not more
pixels, but a what-memory that records hidden change and a training objective that supervises
endpoint persistence''; its scaling evidence is read as showing that no amount of capacity
supplies this. MBench likewise reads its results as a capability gap between visual
plausibility and world modelling. Under this diagnosis the remedy is architectural and
expensive: new memory modules, new training objectives, retraining at scale.

This paper argues that the diagnosis is incomplete, and for an important class of models
simply wrong.

\subsection{The state is already there --- the runtime throws it away}

Our central experiment is a restore. We take a Markovian action-conditioned world model
(Cosmos3), generate a rollout, stop, and snapshot two objects the \emph{runtime} holds: the
current observation, as a \texttt{uint8} array, and the generator's RNG state. We then drive
the model elsewhere --- a genuine excursion, enough to destroy any implicit continuity ---
and afterwards restore the snapshot and continue. The continuation is compared against an
oracle that never left.

It is \textbf{bit-identical}: DINOv2~\cite{dinov2} similarity 1.0, pixel error 0.0. Dropping the RNG from
the snapshot costs 0.017; not restoring at all costs 0.065, roughly four times the RNG term.
A state ablation (none 0.941 / RNG-only 0.935 / observation-only 0.983 / full 1.000) shows
the observation is the load-bearing component and the RNG closes the final gap to exactness.

The consequence is sharp. If the future of a world can be reproduced \emph{exactly} from a
snapshot the serving layer could trivially have taken, then the model did not lack the
capability to continue that world, and no training objective was needed to supply it. What was
missing was that \textbf{nobody saved the state}. The failure the benchmarks measure is, for
this model class, a \emph{runtime} failure wearing the costume of a model failure --- and the
two are distinguishable, by exactly this experiment.

We call this the \textbf{Runtime-Discard Hypothesis} --- that the failure the benchmarks
attribute to the model is, for this class, the runtime discarding recoverable state --- and we
stress what makes it credible: it is established constructively, by restoring and observing
bit-exactness, not inferred from a model's behaviour. An argument of the form ``the model
shows continuity, therefore it must possess memory, therefore the runtime must be discarding
it'' is circular; a bit-exact restore is not. The rest of the paper builds the abstraction
that follows from the hypothesis and, in Section~\ref{sec:eval-c3}, confirms it on a second,
non-Markovian model --- where the recoverable state is a memory bank rather than a frame, and
restore is again bit-exact.

\subsection{Why serving discards it: an inherited assumption}

The discard is not an oversight but a design premise. Modern generative serving is
\textbf{request-centric}: state is allocated to serve a request and reclaimed at its boundary.
This is correct --- indeed optimal --- under an assumption inherited from language-model
serving, that any useful runtime state is \textbf{recomputable on demand}. An LLM KV cache is
an \emph{optimization} over re-running the prefix; it can be evicted freely because it is
always reconstructible from tokens that were kept.

World models violate this assumption. Their runtime state carries a \textbf{non-recomputable
kernel} --- the RNG stream, and in non-Markovian models the memory bank or recurrent
activations --- that no amount of re-running reconstructs, because the sampling noise that
produced a particular world is not a function of the inputs. When such state is reclaimed at a
request boundary, it is not evicted; it is destroyed. A world trajectory spanning many
requests therefore loses the object that defines \emph{where the world is}.

The repair is an abstraction, not a heuristic. The missing unit of work is the
\textbf{session}; the missing object is the \textbf{Persistent Computational State}.

\subsection{A note on terminology}

``Persistence'' is used in two distinct senses in this literature, and we fix ours explicitly.
WRBench uses it for the requirement that an unobserved world keep \emph{evolving} --- the moon
continues its orbit, the cat finishes jumping onto the bed --- so that its failure mode is a
world frozen at the moment observation stopped. We use it in the complementary, computational
sense: the requirement that state which \emph{should} be carried across a request boundary in
fact survives it, so that our failure mode is a world that cannot be recovered at all. The two
are halves of one deficiency --- a system with no state object neither advances what should
advance nor preserves what should be preserved --- and our claim is deliberately the narrower
and more mechanical of the two: not that a runtime can simulate hidden events, but that it can
guarantee that a world it was asked to remember is returned unchanged.

\subsection{Contributions}

We take the benchmarks' finding as established and ask the next question: \emph{what,
precisely, is the lost thing, and whose job is it to keep it?}

\begin{itemize}[leftmargin=1.4em]
\item \textbf{C1 --- Diagnosis.} The failure is a serving gap, established by bit-exact restore
  rather than by inference from behaviour. For a Markovian world model the required state is
  external, explicit, and small, and persistence needs no retraining.
\item \textbf{C2 --- The PCS abstraction.}
  $\mathrm{PCS}(M) = \arg\min_{S}\{\,|S| : D(\restore(S),\oracle) \le \epsilon\,\}$ --- the
  minimal non-recomputable state whose restoration is indistinguishable from never having left
  --- with an API, four invariants, and a layer boundary. This doubles as an \emph{operational
  definition of world state}, a question the roadmap literature poses but does not answer
  measurably.
\item \textbf{C3 --- Discovery by measurement.} A fingerprinting procedure that classifies each
  runtime buffer as recomputable or must-snapshot. It recovers \texttt{\{observation, RNG\}}
  for Cosmos3 and a memory-bank PCS for WorldMem \emph{without model-specific engineering}, and
  its ablation profile --- flat versus gradient --- fingerprints the model's memory
  architecture.
\item \textbf{C4 --- A session-centric runtime, semantically validated.} Return-consistency as
  a \emph{conformance test} for restore and fork --- a criterion of world preservation that
  bit-exactness cannot express, since a restore can be byte-perfect on the wrong state and a
  legitimately re-grounded rollout is not byte-equal to its anchor. The runtime that this test
  validates then yields the evaluation's systems results (Section~\ref{sec:eval}): a planner's
  compute falls from $O(T^2)$ replay to $O(T)$, checkpoint/restore is effectively free, and
  eviction must be relevance-keyed.
\end{itemize}

\paragraph{Scope.} The exactness result is bit-exact on the Markovian model, where the
observation \emph{is} the complete visible state, and again bit-exact on the non-Markovian one
through its memory bank; we do not claim exactness for arbitrary architectures. The
leave-and-return protocol is shared in spirit with MBench's departure-return trajectories and
WRBench's viewpoint intervention, and we claim no novelty for the protocol itself. Our
methodological contributions are elsewhere: three confound controls the benchmarks do not
apply (a generative-drift floor, an attractor control, and a prompt-leakage control,
Section~\ref{sec:eval}), and three scoring corrections without which a conformance measurement
inverts (Section~\ref{sec:conformance}).

\paragraph{Roadmap.} The argument runs in five steps. Section~\ref{sec:gap} shows why the unit of
serving must be a \emph{session} rather than a request. Section~\ref{sec:pcs} defines PCS and
states its contract. Section~\ref{sec:discover} shows the contract is not hand-engineered but
\emph{discovered by measurement}. Section~\ref{sec:runtime} builds the session-centric runtime,
and Section~\ref{sec:conformance} gives the conformance test that decides when a restore is
correct. Section~\ref{sec:eval} evaluates; Sections~\ref{sec:impl}--\ref{sec:conclusion} cover
implementation, related work, and what the abstraction does and does not settle.

\section{The Request-Centric Gap --- why a session?}
\label{sec:gap}

Persistence is unattainable under request-centric serving for reasons that have nothing to do
with the model. The account is short and worth making precisely, because it is what turns a
measured failure into a systems claim.

Let a world trajectory be $\tau = (o_0, a_1, o_1, a_2, o_2, \dots)$, where each observation
$o_t$ is produced by a model $M$ from the action $a_t$ and the model's \textbf{runtime state}
$H_t$ --- the union of everything the next step reads: key-value blocks, latent caches,
recurrent activations, the sampling RNG stream, the last observation. Concretely
$o_t = M(a_t, H_t)$ and $H_{t+1} = \mathrm{update}(H_t, a_t, o_t)$. The trajectory is well
defined only while $H_t$ is carried forward faithfully.

\paragraph{The request-centric claim.} A request-scoped runtime allocates state to serve a
request and reclaims it at the boundary. When the next action arrives as a fresh request, the
runtime reconstructs an \emph{incomplete} $H'_t \subset H_t$ --- whatever is cheap to rebuild,
typically the last observation and a re-prefill, but not the RNG stream and not accumulated
memory. The continuation is conditioned on $H'_t$, so $o_{t+1}$ departs from the true
trajectory, and the departure compounds with horizon. \textbf{Persistence is impossible by
construction.} Not because the model is weak; because the runtime never let it keep what it
needs.

\paragraph{Why this is not merely a restatement of the benchmarks.} The 2026 memory benchmarks
(Section~\ref{sec:related}) measured the \emph{symptom} at scale and attributed it to the
model. The account above predicts the same symptom from the runtime alone, and the two
explanations are distinguishable by a single experiment: if the state is genuinely absent from
the model, no snapshot can restore the trajectory; if the runtime discarded it, a snapshot can.
Section~\ref{sec:eval-c3} runs that experiment. Restoring the runtime's non-recomputable state
--- a \texttt{uint8} observation, a memory bank, or a windowed KV context, plus the
generator state --- reproduces the never-left continuation byte-identically on three models with
nothing architectural in common. The runtime explanation survives; the model-deficit
explanation does not, at least for this class.

\paragraph{The session claim.} A \emph{session} is the abstraction that \textbf{owns} $H_t$
across requests. If it preserves that state the trajectory stays well defined. This immediately
raises the question the rest of the paper answers, because $H_t$ is large and mostly
rebuildable: \textbf{which subset must actually persist?} That subset is PCS
(Section~\ref{sec:pcs}), and it is discovered rather than assumed (Section~\ref{sec:discover}).

\begin{figure}[tbp]
\centering
\includegraphics[width=\linewidth]{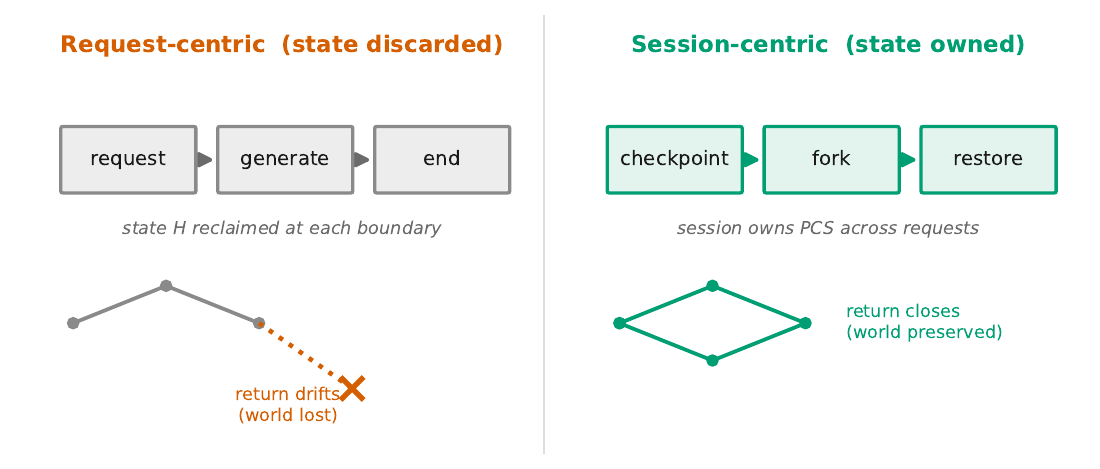}
\caption{\textbf{Request-centric vs.\ session-centric.} \emph{Left:}
\texttt{request $\to$ generate $\to$ end}; state is discarded at the boundary and the
trajectory breaks on return. \emph{Right:}
\texttt{session $\to$ checkpoint $\to$ fork $\to$ restore}; state is owned across requests and
the trajectory survives.}
\label{fig:session}
\end{figure}

\section{Persistent Computational State --- definition and contract}
\label{sec:pcs}

\begin{quote}
\textbf{Persistent Computational State (PCS) is the minimal non-recomputable computational
state that must survive across requests to guarantee semantic continuation of a generative
session.}
\end{quote}

Made precise: let $H$ be the model's full runtime state; $\contn(\restore(S))$ the
continuation after restoring a subset $S \subseteq H$; $\oracle$ the never-left continuation;
$D$ a semantic distance (the return-consistency metric of Section~\ref{sec:conformance});
$\epsilon$ a tolerance, with $\epsilon = 0$ meaning bit-exact. Then

\begin{definition}[PCS]
$\mathrm{PCS}(M) = \arg\min_{S \subseteq H} |S| \ \text{ s.t. } \
D(\contn(\restore(S)), \oracle) \le \epsilon.$
\end{definition}

Three properties follow.

\begin{itemize}[leftmargin=1.4em]
\item \textbf{Minimality.} No proper subset suffices; every element is load-bearing.
\item \textbf{Recoverability.} Restoring PCS yields a continuation within $\epsilon$ of never
  leaving. The complement $H \setminus \mathrm{PCS}$ is \emph{derivable} --- recomputable from
  PCS by re-running compute --- and need not be stored.
\item \textbf{Model-dependence.} $\mathrm{PCS}(M)$ varies by model, and the variation is not
  cosmetic: a flat \texttt{\{observation, RNG\}} for a Markovian model versus a set localised
  to an internal memory mechanism for a non-Markovian one. It is measured, not assumed
  (Section~\ref{sec:discover}).
\end{itemize}

\paragraph{The partition, and where the LLM intuition breaks.} PCS induces
$H = \mathrm{PCS} \uplus \mathrm{Derivable}$, where a component is derivable iff it
reconstructs from PCS by re-running compute. This is precisely where the mental model behind
prefix caching fails for world models: the RNG stream and the recurrent or memory kernel are
\textbf{non-derivable}, so a runtime that ``caches by recompute'' is not inefficient here but
\emph{incorrect}, and silently --- it returns a plausible world rather than the one that was
left.

\paragraph{The contract.} A definition does not tell a systems reader what is guaranteed, so
PCS is specified as a contract: an API, its invariants, and a boundary. The API is four
operations over a session $s$ and opaque handles $c$:
\[
c = \texttt{checkpoint}(s) \quad\cdot\quad s' = \texttt{restore}(c) \quad\cdot\quad
c' = \texttt{fork}(c) \quad\cdot\quad \texttt{delete}(c)
\]
Two states are \textbf{semantically equivalent} ($s_1 \approx s_2$) iff
$D(\contn(s_1), \contn(s_2)) \le \epsilon$ --- equivalence with respect to the
return-consistency metric of Section~\ref{sec:conformance}, \emph{not} byte identity. The
runtime guarantees:

\begin{itemize}[leftmargin=1.4em]
\item \textbf{(I1) Round-trip.} $\restore(\ckpt(s)) \approx s$. \emph{Measured bit-exact on
  both model classes: DINOv2 1.0, pixel error 0.0 (Section~\ref{sec:eval-c3}).}
\item \textbf{(I2) Durability.} A checkpoint is an immutable snapshot, not a live alias;
  operations issued between \texttt{checkpoint} and \texttt{restore} do not affect it, and a
  session survives the death of the process that created it. \emph{Measured within a process:
  restore survives another session's rollout running in between. Measured across a process
  boundary on \textbf{both} models: an oracle, a checkpoint-then-die process, and a fresh resume
  process --- sharing only a file on disk --- return byte-identically (DINOv2 1.0, pixel error
  0.0), durability in its strongest form (Section~\ref{sec:eval-c3}).}
\item \textbf{(I3) Fork independence.} A child's continuation is semantically independent of the
  parent's. The mechanism realises this by copy-on-write; the contract requires only semantic
  independence, which the conformance test certifies.
\item \textbf{(I4) Snapshot completeness.} \texttt{checkpoint} captures exactly PCS and nothing
  derivable. This is the invariant that makes the bound $|\ckpt| = |\mathrm{PCS}|$ rather than
  $|H|$.
\end{itemize}

\paragraph{Boundary.} PCS is a \emph{semantic} classification layer. It decides \emph{which}
buffers are non-recomputable and must be captured (Section~\ref{sec:discover}) and \emph{when}
to re-ground or evict them (Section~\ref{sec:runtime}); the lower runtime owns paged
allocation, copy-on-write and placement; storage owns serialization; the framework owns model
code and the RNG source. The line is drawn at derivability --- a buffer is PCS iff it is
non-recomputable for this model under this probe, regardless of which layer physically holds
it. That is what separates PCS from a checkpoint facility (which stores bytes without asking)
and from a KV cache (which stores only what \emph{is} recomputable); see
Section~\ref{sec:related}.

\paragraph{A distinction the implementation forced.} Not everything a runtime holds about a
session is state. Our adapters retain the last rendered frame for scoring, but the
non-Markovian model's \texttt{restore} never reads it --- the continuation is rebuilt from the
bank. Counting it as PCS would have reported 2.95~MB per session against a true 185~KB,
inflating the host cost of the whole design by an order of magnitude. The contract therefore
distinguishes \emph{state} from \emph{witness}, with a mechanical test: \textbf{a buffer is PCS
iff \texttt{restore} reads it.} For the Markovian model the observation passes that test and is
PCS; for the non-Markovian one it does not and is witness. The same buffer, opposite
classifications, decided by behaviour rather than by name.

\section{Discovering PCS by measurement}
\label{sec:discover}

To manage PCS a runtime must know what it \emph{is} for a given model. Prior practice offers
two unsatisfying answers: \textbf{save everything}, which is correct but carries unbounded,
horizon-scaling memory and defeats the purpose; or \textbf{hand-engineer} a per-model state
list, which is compact but fragile, unverified, and re-derived by hand for every model. Neither
yields the minimal set and neither ports. We replace both with a procedure.

Call a model $M$ \textbf{fingerprintable} if it exposes three things and nothing
architecture-specific beyond them: (i) runtime state decomposing into addressable components
$H = \{h_1, \dots, h_k\}$; (ii) \texttt{checkpoint}$(S)$ / \texttt{restore}$(S)$ over any
subset; and (iii) a return-consistency metric $D$ with tolerance $\epsilon$
(Section~\ref{sec:conformance}) plus a \emph{probe} --- a distribution of leave-and-return
trajectories.

\begin{quote}
\textbf{Procedure} \textsc{Fingerprint}$(M, D, \epsilon, \text{probe})$\textbf{:}
\begin{enumerate}[leftmargin=1.6em,label=\arabic*.]
\item \textbf{(necessity)} for each $h_i \in H$: snapshot $H \setminus \{h_i\}$, run the probe,
  measure $D$; label $h_i$ \emph{necessary} if $D > \epsilon$.
\item \textbf{(sufficiency)} let $N$ be the necessary set; snapshot exactly $N$ and run the
  probe. If $D \le \epsilon$, return $\mathrm{PCS} := N$.
\item \textbf{(redundancy fallback)} otherwise --- correlated or mutually-recomputable buffers,
  where no single one is individually necessary --- greedily add the component that most reduces
  $D$ until $D \le \epsilon$.
\item \textbf{return} $\mathrm{PCS}$ and $\mathrm{Derivable} := H \setminus \mathrm{PCS}$.
\end{enumerate}
\end{quote}

The returned set \emph{is} the runtime's snapshot plan.

\paragraph{Why this is a procedure and not a per-model trick.} \textsc{Fingerprint} assumes
nothing about architecture, modality or training objective --- only that the model can be
checkpointed and scored. It is the same loop for a diffusion model and for a recurrent one; the
\textbf{output} varies, the \textbf{procedure} does not. Applied to the three models here it
returns \texttt{\{observation, RNG\}} (1.38~MB, flat, constant in horizon),
\texttt{\{memory bank, pose, RNG\}} (185~KB, growing at roughly 37~KB per retained frame), and
\texttt{\{KV context, position indices, RNG\}} ($\approx$1.67~GB, large but windowed --- fixed by
the local-attention span) --- sets that share no component beyond the RNG and differ in the
property that matters most for a serving layer, which is whether and how they grow.

\paragraph{On necessity tests that pass for the wrong reason.} Step 1 asks whether dropping a
component degrades the continuation. A component can appear unnecessary simply because the model
never read the snapshot at all, in which case every subset ``passes''. The procedure is
therefore run with a control: replace a component with a \emph{wrong} value rather than removing
it, and require the continuation to degrade. On all three models, corrupting only the RNG moves the
continuation off bit-exactness (DINOv2 to 0.979 and 0.970 on the first two; latent agreement to
$3\%$ of elements exact on Matrix-Game 2.0), which is what establishes that the
snapshot is load-bearing rather than ignored. Without that control,
Section~\ref{sec:eval-c3}'s headline would be unfalsifiable.

\begin{figure}[tbp]
\centering
\includegraphics[width=\linewidth]{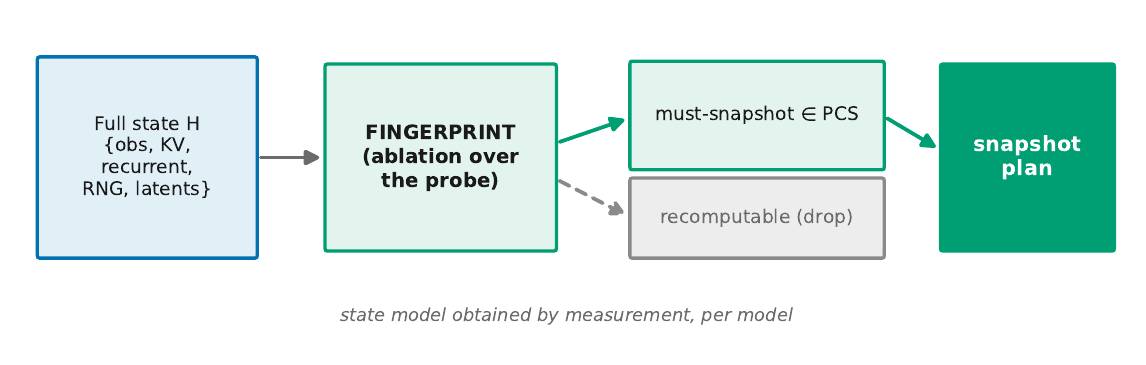}
\caption{\textbf{PCS discovery by measurement.} The full runtime state $H$ is passed through
\textsc{Fingerprint} (ablation over the probe); each component is classified must-snapshot
($\in$ PCS) or recomputable (drop). The returned set is the runtime's snapshot plan ---
obtained per model, by measurement, not asserted.}
\label{fig:discovery}
\end{figure}

\section{A session-centric runtime over PCS}
\label{sec:runtime}

The mechanism is the community's; the semantic layer is ours, and it is deliberately thin.

\paragraph{Unit = session.} The runtime exposes \texttt{checkpoint / restore / fork / retrieve}
over \textbf{PCS}, not over requests. Its contribution is \emph{what these operate on} and
\emph{how a restore is validated} (Section~\ref{sec:conformance}) --- not a new allocator or
memory pool.

\paragraph{Structure.} The GPU is time-multiplexed: one session holds the model, every other
session's state lives on the host as PCS. The engine is therefore a single-server queue, which
is what makes device memory constant in the number of sessions while host memory grows linearly
(Section~\ref{sec:eval-scaling}). This is a scoping decision, not a limitation to apologise for
--- a concurrent multi-GPU runtime is a different system and its numbers would not be comparable
to these --- but it does mean that ``$N$ sessions'' throughout this paper denotes $N$
\emph{resident contexts}, not $N$ concurrent rollouts.

\paragraph{The two semantic decisions the runtime owns.} \emph{(1) What to snapshot} --- the PCS
partition from Section~\ref{sec:discover}, measured per model rather than assumed. \emph{(2)
What to evict, and when} --- a session's \textbf{persistence horizon} bounds how long a rollout
may run before it must be re-grounded, and the same notion drives eviction: keep the state that
still contributes to a return, drop the rest. For a pose-keyed memory the frames that matter on
return are the \textbf{anchor-side} ones, not the recent ones, so the policy is
\textbf{relevance-keyed --- the inverse of the recency eviction that LLM serving uses.}
Section~\ref{sec:eval-evict} measures the gap: at a tight budget, relevance-keyed retention
holds sixteen worlds of sixteen while recency-keyed holds six.

\paragraph{On scheduling --- what we tried, and why the paper does not claim it.} An earlier
version of this work deferred scheduling as future work. We instead built it and report the
result, because the negative outcome bounds the claim more usefully than a promise would. Two
predictive designs were implemented and measured (Section~\ref{sec:eval-neg}). \emph{Predictive
eviction} --- shrink while slack remains, to stay out of the regime where deletion is the only
move --- is monotonically \emph{worse} than acting late, because the runtime escalates shrink
$\to$ nothing-left-to-shrink $\to$ drop, so a lower trigger pushes more sessions to the bottom
of their range. \emph{Predictive admission} --- project demand and refuse what cannot be kept
whole --- refuses worlds that would have completed, because a realistic arrival trace peaks near
six live sessions and never oversubscribes the budget. On a persistent-world load that \emph{does}
oversubscribe (Section~\ref{sec:eval-oversub}, peak concurrency 29), both still fail: predictive
eviction is byte-for-byte identical to reactive because the retention floor, not the trigger time,
bounds what eviction can save; predictive admission only helps past the point where the system has
already collapsed (Section~\ref{sec:eval-neg}). The runtime therefore claims a \emph{mechanism} and
a \emph{policy for what to keep}, and explicitly does not claim a scheduler.

\begin{figure}[tbp]
\centering
\includegraphics[width=0.92\linewidth]{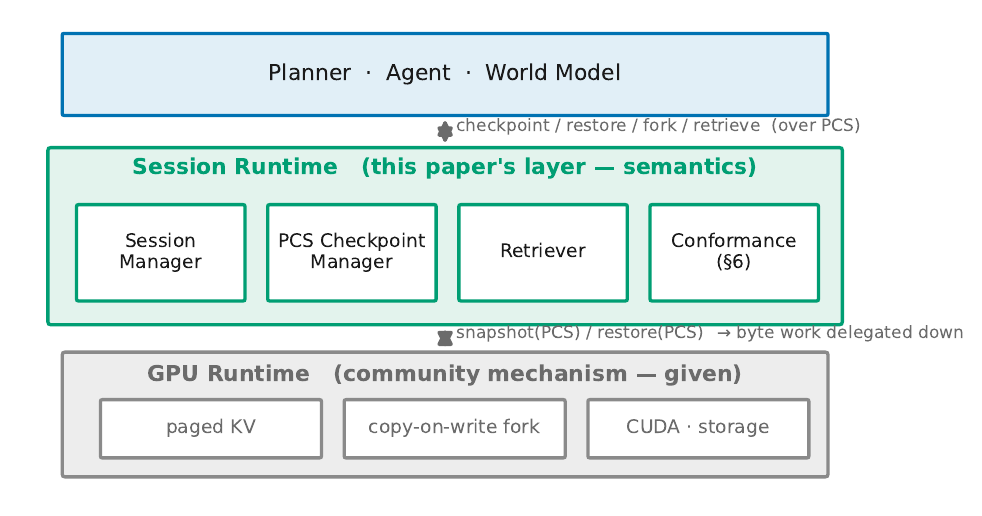}
\caption{\textbf{Runtime architecture.} \texttt{Planner / Agent / World Model} above; the
\textbf{Session Runtime} (session table, PCS checkpoint manager, retriever, conformance) is
this paper's layer; the \texttt{GPU runtime} (paged KV, copy-on-write fork, CUDA, storage) is
community mechanism, consumed as given. Byte work is delegated downward.}
\label{fig:arch}
\end{figure}

\section{Return-consistency as the conformance test}
\label{sec:conformance}

\paragraph{Claim.} A restore or fork is \emph{correct} iff the continuation \textbf{preserves
the world}, and this is a property \textbf{bit-exactness cannot express}, in both directions:

\begin{itemize}[leftmargin=1.4em]
\item \textbf{Lossy but correct.} Re-grounding a rollout that cannot return bit-exactly ---
  after a long excursion, say --- is still correct if the world is preserved. A bit-exact
  criterion would wrongly reject it.
\item \textbf{Byte-successful but memory-lost.} A fork that copies bytes successfully while
  dropping a non-derivable object silently loses memory. A bit-exact check \emph{of the wrong
  state} passes it; return-consistency fails it. This is the ``KV-only fork corrupts
  non-KV-state models'' hazard, turned into a test.
\end{itemize}

\paragraph{The test.}
\texttt{checkpoint $\to$ leave $\to$ restore(PCS) $\to$ continue $\to$ return-consistency $>$
floor.} It rejects the memoryless baseline, which cannot return, and it rejects the degenerate
``always show the anchor'' attractor --- a forward-motion control catches the latter, since a
genuine return must also move correctly rather than merely display the start frame.

\paragraph{Implementing it is where the subtlety is.} Three ways of scoring this test produce
confidently wrong answers, and we hit all three before arriving at the current form.

\emph{Score at the return, not after the run.} A session that finishes releases its snapshots.
Scoring post hoc therefore marks every cleanly-completed session as having lost its anchor ---
inverting the result exactly. Conformance is a property of the moment of return and must be
recorded then.

\emph{A lost anchor is a failure, not a missing sample.} Scoring only the sessions whose anchor
survived is survivorship bias of the most damaging kind here: the policy that destroys the most
anchors reports the cleanest mean. In one early run sixteen sessions yielded a single score of
0.9996. A lost world scores zero.

\emph{The pressure must be caused, not arranged.} Eviction only touches suspended sessions, so a
workload in which each session runs its legs back to back never exposes one to eviction during
its own excursion. Every policy then looks identical. The situation has to arise from arrival
rate and think time rather than from the order in which the harness happened to submit.

\begin{figure}[tbp]
\centering
\includegraphics[width=\linewidth]{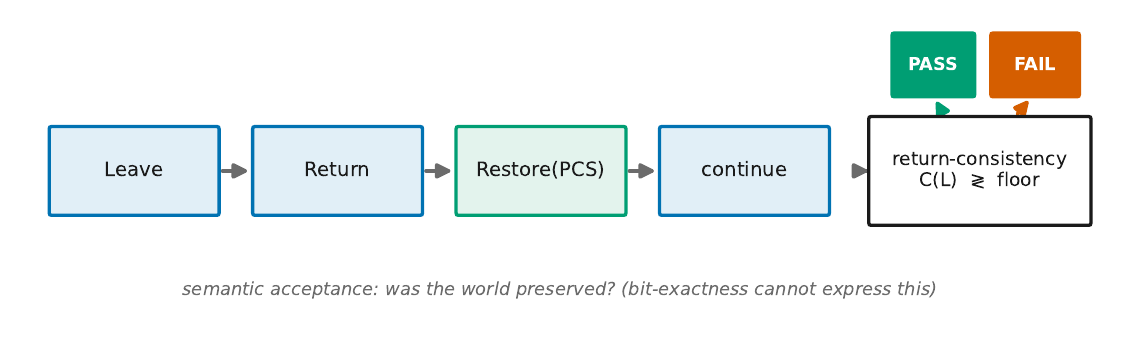}
\caption{\textbf{Conformance.}
\texttt{Leave $\to$ Return $\to$ Restore(PCS) $\to$ continue}, then return-consistency $C(L)$
against a floor $\Rightarrow$ pass / fail. Semantic acceptance: was the world preserved?
Bit-exactness cannot express this.}
\label{fig:conformance}
\end{figure}

\section{Evaluation}
\label{sec:eval}

\subsection{Setup}

One A800-80GB. Three models spanning the memory architectures the abstraction has to cross:
\begin{itemize}[leftmargin=1.4em]
\item \textbf{Cosmos3-Nano}~\cite{cosmos} --- Markovian, action-conditioned diffusion; conditions each chunk
  on a single observation.
\item \textbf{WorldMem}~\cite{worldmem} --- non-Markovian; carries an explicit memory bank (five arrays: latent
  frames, actions, poses, camera-to-world, frame indices) with state-aware memory attention.
\item \textbf{Matrix-Game 2.0}~\cite{matrixgame2} --- autoregressive causal-diffusion world model with an
  addressable memory channel: its runtime state is a windowed KV context (per-block key/value
  caches for the spatial, mouse, and keyboard streams plus a cross-attention cache), each block
  tagged with explicit position indices, capped by a local-attention window. Driven by a
  scripted camera leave-and-return (yaw away, then yaw back).
\end{itemize}

The first two are driven through the same \texttt{SessionRuntime}: one session resident on the device at
a time, every other session's state held as PCS on the host. The workload is leave-and-return
--- anchor, excursion, return --- issued as three requests per session and interleaved across
sessions, so a session is suspended (and therefore evictable) \emph{during} its own excursion
rather than running to completion first.

Conformance is scored \textbf{at the moment of return}, not after the run. A session that closes
releases its snapshots; scoring post hoc would mark every cleanly-completed session as having
lost its anchor. This is not a detail --- it inverted the eviction ranking in an early version
of the harness.

\subsection{One procedure, two shapes of state (C3)}
\label{sec:eval-c3}

The load-bearing claim is not that a snapshot can be restored --- that is a checkpoint API. It
is that \emph{one measurement procedure}, given no model-specific input, arrives at the minimal
restorable state for models whose internals have nothing in common, and that restoring it is
exact in both.

\begin{table}[t]
\centering
\caption{PCS discovered by the same procedure on three model classes.}
\label{tab:pcs}
\begin{tabular}{lccc}
\toprule
 & \textbf{Cosmos3} & \textbf{WorldMem} & \textbf{Matrix-Game 2.0} \\
\midrule
profile & flat & gradient & windowed \\
memory architecture & observation & growing bank & addressable KV context \\
PCS components & \texttt{\{obs 1{,}382{,}400\,B,} & \texttt{\{bank 185{,}280\,B,} &
  \texttt{\{kv 973\,MB $+$ 649\,MB,} \\
 & \texttt{rng 8\,B\}} & \texttt{pose 24\,B, rng 8\,B\}} &
  \texttt{x-attn 47\,MB, pos-idx, rng\}} \\
PCS total & 1{,}382{,}408\,B & 185{,}312\,B & $\approx$1.67\,GB \\
restore vs.\ never-left (DINOv2) & \textbf{1.0} & \textbf{1.0} & \emph{latent-exact}$^\dagger$ \\
restore vs.\ never-left (pixel) & \textbf{0.0} & \textbf{0.0} & \emph{latent-exact}$^\dagger$ \\
byte-identical continuation & \textbf{yes} & \textbf{yes} & \textbf{yes} \\
continuation with the RNG replaced & 0.9786 & 0.9704 & 3\% latent-exact$^\dagger$ \\
cross-process byte-identical & \textbf{yes} & \textbf{yes} & \textbf{yes} \\
\bottomrule
\end{tabular}

\smallskip
{\footnotesize $^\dagger$ Matrix-Game 2.0 is compared in \emph{latent} space (the model's
native output before VAE decode). Capture$\to$restore gives max absolute latent difference
$0.0$ (\texttt{torch.equal} true) --- byte-identical, strictly stronger than pixel equality,
which follows because the VAE decode is deterministic. Replacing the RNG collapses latent
agreement to $3\%$ of elements exact (max $|\Delta|$ $3.875$), confirming the snapshot is
load-bearing.}
\end{table}

The last row matters more than it looks. A restore test passes trivially if the model never
reads the snapshot; corrupting only the RNG and observing that the continuation \emph{degrades}
is what shows the snapshot is actually load-bearing. Without it, row four proves nothing.

Two things worth stating because a reader will assume the opposite:

\paragraph{Size does not track richness; growth is the axis that matters.} Cosmos3's state is a
full $480\times960$ \texttt{uint8} observation (1.38~MB); WorldMem's is a compressed latent bank
(185~KB) --- $7.5\times$ smaller despite being the ``richer'' model; Matrix-Game 2.0's is a
windowed KV context three orders of magnitude larger again ($\approx$1.67~GB, dominated by
973~MB of spatial and 649~MB of action key/value tensors across 30 blocks). The three span
$185\,\text{KB} \to 1.38\,\text{MB} \to 1.67\,\text{GB}$, and the ordering does not follow any
intuition about model capability. The distinction that matters for a serving layer is not size
but \emph{growth under horizon}: WorldMem's bank grows unboundedly ($\approx$37~KB per retained
frame), Cosmos3's observation is constant, and Matrix-Game's KV context is large but
\emph{fixed} --- capped by the local-attention window (\texttt{local\_attn\_size}), so it does
not grow with rollout length. A host budget must be sized against the second fact, not the
first; the three profiles --- constant, growing, and windowed --- are exactly the three the
fingerprint returns.

\paragraph{\texttt{uint8}, deliberately.} The claim being tested is that a \emph{\texttt{uint8}}
observation plus the generator state restores exactly. Snapshotting \texttt{float32} would make
the claim easier than the one measured, so the adapter quantises before storing.

\paragraph{Cross-check.} The bank component, 185{,}280~B, is exactly the checkpoint size
measured independently for the same horizon by the standalone script
(\texttt{worldmem\_ckpt\_size.json}, $K{=}4$). The runtime reaches it through an entirely
different path --- the bank is sliced into per-frame records and re-stacked on restore --- and
lands on the same byte count.

\paragraph{Across a process boundary.} I2's strongest form is durability across process death,
not merely across intervening work. We run it as three separate process invocations wired so the
kill is real: an \emph{oracle} process runs anchor\,$\to$\,away\,$\to$\,return and never dies; a
\emph{checkpoint} process runs anchor\,$\to$\,away in a fresh interpreter, writes the PCS to
disk, and exits (and is then killed); a \emph{resume} process --- a brand-new interpreter that
re-initializes CUDA, reloads weights, and re-selects kernels --- loads the PCS from that file,
restores, runs the same return leg, and compares to the oracle. On \textbf{all three} models the
resumed continuation is \textbf{byte-identical} across the boundary
(\texttt{bit\_identical\_across\_process = true}): WorldMem restoring from a $\sim$260~KB bank file
and Cosmos3 from a 1.38~MB observation file (DINOv2 1.0, pixel error 0.0), and Matrix-Game 2.0
from a 1.68~GB KV-context checkpoint written to disk by the dying process --- its restored return
leg matching the oracle at max absolute latent difference $0.0$ (\texttt{torch.equal} true), the
strongest of the three since it holds at the model's native latent resolution. Each sharing
nothing with the checkpoint process but that file. This is the direct evidence that a session
outlives the process that made it, not merely a request within one. The anticipated weaker outcome
--- semantically exact but not byte-exact, from cross-process kernel selection --- did not occur on
any model; durability holds in its strongest form across all three classes. That byte-exactness
survives a full CUDA re-initialization does not
narrow the conformance test of Section~\ref{sec:conformance}, which stays semantic by design: it
does not \emph{need} byte-exactness and would still accept a legitimately re-grounded restore that
lacked it.

\subsection{Session scaling}
\label{sec:eval-scaling}

\begin{table}[t]
\centering
\caption{$S$ sessions, one leave-and-return each, WorldMem, real clock.}
\label{tab:scaling}
\setlength{\tabcolsep}{4pt}
\resizebox{\textwidth}{!}{%
\begin{tabular}{rrrrrrrrrrrr}
\toprule
$S$ & wall (s) & steps & s/step & p95 lat (s) & p95 wait (s) & ckpt & restore & switches &
  device (GB) & peak host (MB) & ret-consist \\
\midrule
1  & 37.3  & 20  & 1.864 & 37.4  & 22.3  & 5   & 1  & 0  & 4.714 & 0.964  & 0.9427 \\
2  & 76.2  & 40  & 1.906 & 76.3  & 60.9  & 15  & 6  & 5  & 4.714 & 1.927  & 0.9503 \\
4  & 167.3 & 80  & 2.092 & 167.4 & 150.0 & 31  & 12 & 11 & 4.714 & 3.854  & 0.9484 \\
8  & 324.5 & 160 & 2.028 & 309.3 & 294.0 & 63  & 24 & 23 & 4.714 & 7.708  & 0.9535 \\
16 & 602.5 & 320 & 1.883 & 571.8 & 556.5 & 127 & 48 & 47 & 4.714 & 15.416 & 0.9531 \\
\bottomrule
\end{tabular}}
\end{table}

Three readings.

\paragraph{Device memory is constant; host memory is linear.} 4.714~GB at every $S$ --- the
model and its working set, nothing else. Host cost is 0.0364~MB per session, flat to five
decimal places. This is the direct consequence of the design: the device holds one session, the
rest are PCS on the host. An earlier standalone run reported this shape from a single $S{=}6$
point plus extrapolation; here it is five measured points.

\paragraph{The runtime adds no per-session cost.} Time per generation step is 1.86--2.09~s
independent of $S$. The queue wait grows linearly, as a single-server queue must; that is
scheduling, not overhead.

\paragraph{Sessions is residency, not concurrency.} ``16 sessions'' means sixteen resident
contexts with one rollout in flight, because the GPU is time-multiplexed. Every axis in this
paper is labelled that way. A reader who assumes sixteen concurrent rollouts will read the
constant device memory as impossible rather than as the point.

\paragraph{Measured to 1{,}024 sessions, not extrapolated.} The five real-clock points above stop
at $S{=}16$ because a real rollout at each of them costs a full generation pass; the memory
\emph{structure}, however, is a property of the runtime, not the model, and we sweep it directly.
Driving the runtime to $S{=}1024$ resident sessions holds \textbf{device memory constant at
4.714~GB} across every one of the ten doublings from 1 to 1024, with host memory strictly linear at
0.0373~MB per resident session (38.2~MB at 1024). A companion sweep with mock generation steps
(so that returns are actually scored) reaches $S{=}256$ with the same constant device figure, host
peaking at 247~MB, and \textbf{every session's anchor kept --- return-consistency 1.0, zero anchors
lost, at 100 and at 256 sessions alike}. This turns the counterfactual's device-constant claim from
a 16-point extrapolation into a measured curve two orders of magnitude further out, and it is the
direct check on the projected ceiling: a device-bound runtime collapses at 108 sessions
(Table~\ref{tab:counterfactual}), where this one is still flat.

\paragraph{What the alternative would cost.} A flat line is only informative next to one that is
not, so we state the counterfactual explicitly rather than leaving the reader to supply it.
Externalizing state is what makes \emph{suspension} possible: without a snapshot a session
cannot be put down and picked up, so every open session must stay resident and device memory
grows with the session count. Over the same two measured constants --- the model at 4.714~GB and
one active rollout's working set at 0.696~GB (the $S{=}6$ peak minus the model, with exactly one
rollout in flight) --- the two regimes are those of Table~\ref{tab:counterfactual}, a factor of
roughly 2{,}300 in the number of sessions the same GPU can carry.

\begin{table}[t]
\centering
\caption{Counterfactual: device-memory ceiling with and without PCS.}
\label{tab:counterfactual}
\begin{tabular}{lll}
\toprule
 & device memory & ceiling on this hardware \\
\midrule
with PCS & \texttt{model} --- constant in $S$ & host-bound: $\approx$247{,}000 sessions at a
  64\,GB host budget \\
without PCS & \texttt{model + }$S\times$\texttt{ working set} & device-bound: $\approx$\textbf{108}
  sessions at 80\,GB \\
\bottomrule
\end{tabular}
\end{table}

\begin{figure}[tbp]
\centering
\includegraphics[width=0.82\linewidth]{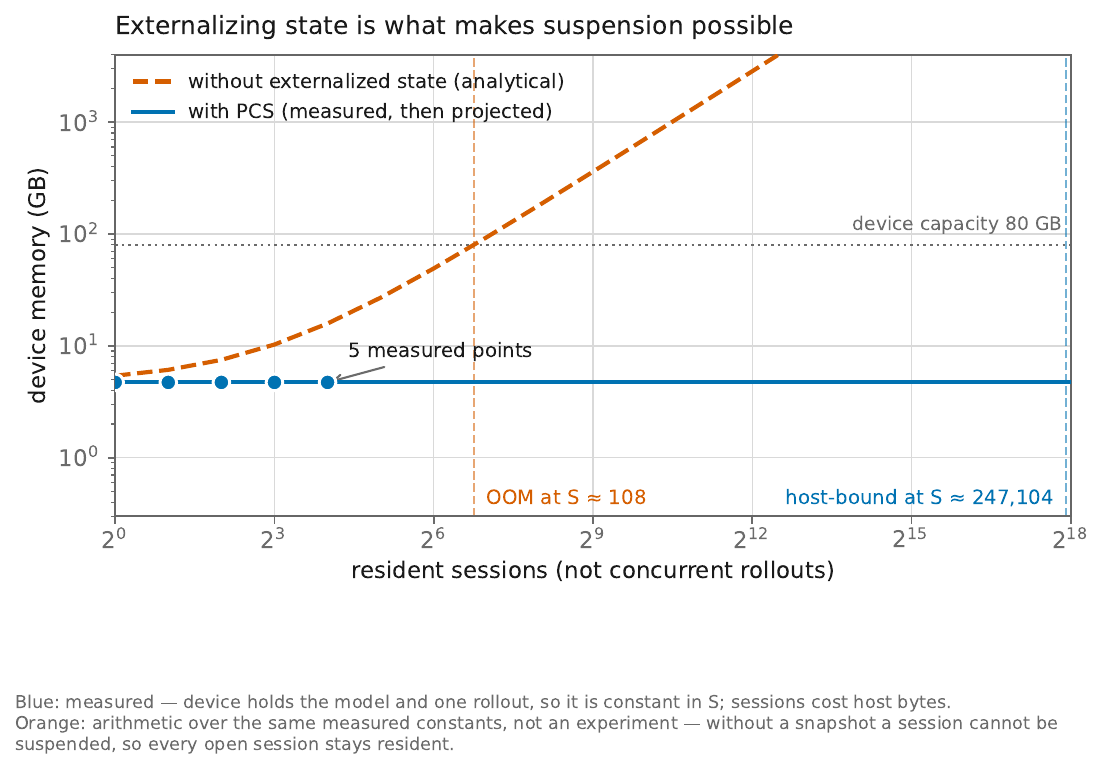}
\caption{\textbf{Sessions per GPU, with vs.\ without PCS.} Blue is measured (device memory
constant in $S$); orange is derived from two measured constants (device-bound, growing with
$S$). Externalizing state is what makes suspension possible; the gap is $\approx$2{,}300$\times$
in resident sessions.}
\label{fig:counterfactual}
\end{figure}

\paragraph{This line is arithmetic, and the figure says so.} The measured curve
(Figure~\ref{fig:counterfactual}, blue) is real; the orange one is derived from measured
constants, not run. Measuring it honestly would mean building the concurrent multi-session
runtime this paper deliberately does not build (Section~\ref{sec:runtime}), and reporting a
projection as an experiment is exactly the move that would make the rest of the numbers less
trustworthy. We label it and leave it labelled.

\paragraph{A note on what this argument is not.} It is tempting to frame the saving as
share-and-append forking versus naive copying --- ref-counted blocks whose copy-on-write copy
leg never fires. That framing does not apply here. WorldMem consumes its bank in place, so a
fork \emph{must} copy the PCS and sharing is unavailable for this model class; the comparison
has no two ends. The saving comes one level up, from being able to suspend at all, and the
per-fork cost is bounded by $|\mathrm{PCS}|$ rather than eliminated.

\subsection{What state management costs}
\label{sec:eval-cost}

\begin{table}[t]
\centering
\caption{The ratio the design turns on (median over the $S{=}16$ run).}
\label{tab:cost}
\begin{tabular}{lrr}
\toprule
operation & cost & relative to one generation step \\
\midrule
checkpoint & 0.012~ms & 1 / 154{,}298 \\
restore & 0.012~ms & 1 / 154{,}298 \\
restore vs.\ the $K$-step replay it replaces & --- & 1 / 1{,}173{,}031 \\
generation step & 1.852~s & 1 \\
\bottomrule
\end{tabular}
\end{table}

Aggregated over the $S{=}16$ run: 127 checkpoints and 48 restores total \textbf{2.1~ms out of
602.5~s --- 0.00035\%}. Persistence is not a resource to be budgeted against generation; it is
free at this scale, and the interesting question is therefore not \emph{whether} to snapshot but
\emph{what}.

\subsection{Eviction: which bytes survive decides whether the world does}
\label{sec:eval-evict}

When the host budget binds, every policy prunes each session's bank to the same fraction (12\%,
the retention level the standalone eviction study measured). They free identical bytes and
differ only in \emph{which frames they rank as worth keeping}. That is what makes the comparison
fair.

\begin{table}[t]
\centering
\caption{Return-consistency by host budget, $S{=}16$, WorldMem.}
\label{tab:evict}
\begin{tabular}{lccccc}
\toprule
host budget & anchor & lru & fifo & pcs-aware & anchor $-$ lru \\
\midrule
2\,MB  & 0.6227 & 0.2127 & 0.1186 & 0.7756 & \textbf{+0.4100} \\
4\,MB  & 0.6227 & 0.4110 & 0.2981 & 0.9531 & \textbf{+0.2117} \\
8\,MB  & 0.7415 & 0.6975 & 0.5362 & 0.9531 & \textbf{+0.0440} \\
16\,MB & 0.9531 & 0.9531 & 0.9531 & 0.9531 & 0.0000 \\
\bottomrule
\end{tabular}
\end{table}

\paragraph{The advantage is a function of pressure, not a constant.} At 16~MB no policy evicts at
all (\texttt{shrinks = drops = 0} for all four): the unconstrained footprint is 15.4~MB, so the
budget stops binding and the ranking is never consulted. That row is a measured boundary, not an
extrapolation. As the budget tightens the policies separate monotonically, and at 2~MB
relevance-keyed retention holds 16 of 16 worlds while recency holds 6.

\paragraph{Relevance costs more eviction work, not less.} At 2~MB the anchor policy performs 45
shrinks and \textbf{zero} drops; recency performs 23 shrinks and \textbf{20 drops}, destroying
ten sessions outright. Doing more of the cheap operation is how it avoids the expensive one.

\begin{figure}[tbp]
\centering
\includegraphics[width=\linewidth]{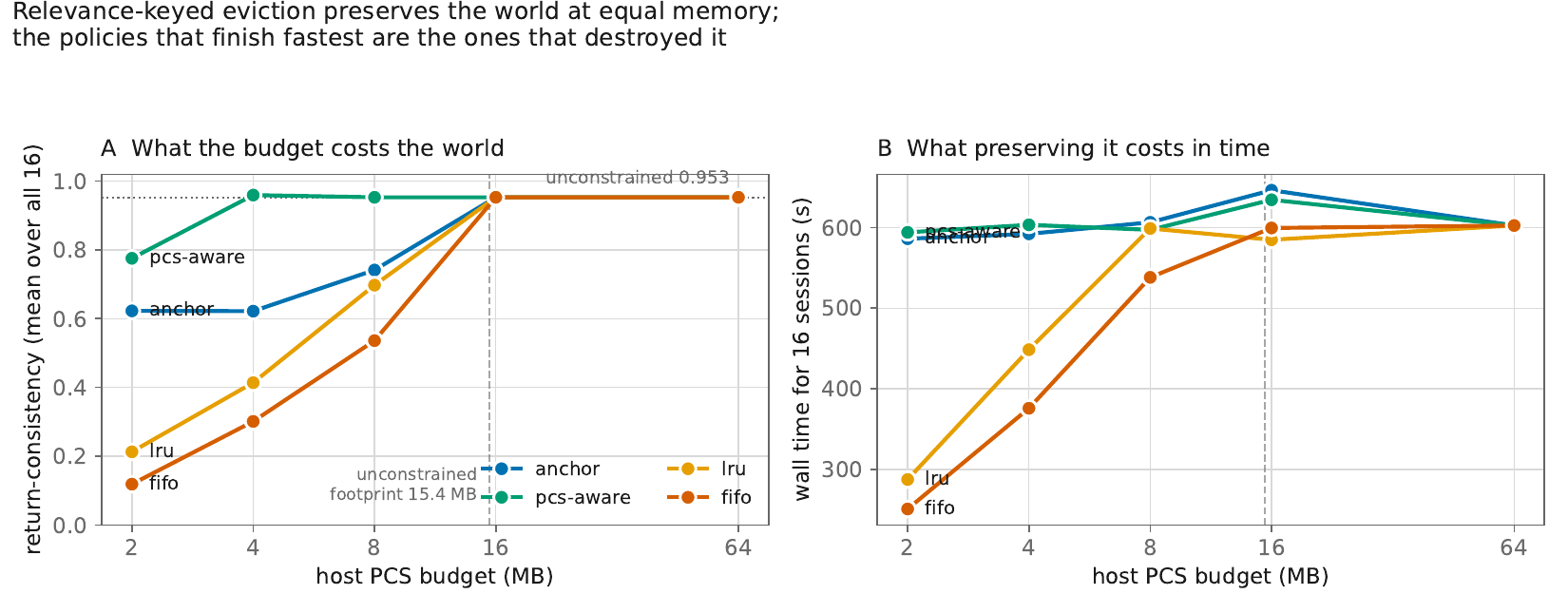}
\caption{\textbf{Eviction cost curve.} Panel A: return-consistency vs.\ host budget
(Table~\ref{tab:evict}) --- relevance and recency separate as pressure rises and the gap widens
to $+0.41$ at 2\,MB. Panel B: wall time, which ranks the policies backwards because a policy that
discards more state generates faster.}
\label{fig:costcurve}
\end{figure}

\paragraph{The faster policies are the ones that destroyed the most.} Wall time at 2~MB: fifo
251~s, lru 288~s, anchor 586~s --- and the unconstrained run is 602~s. A policy that discards
state generates faster, because the bank it attends over is smaller. Reporting latency alone
would rank the policies exactly backwards, which is why the second panel of
Figure~\ref{fig:costcurve} exists.

\paragraph{Why anchor scores identically at 2~MB and 4~MB.} Note that the anchor policy gives the
same 0.6227 at both budgets. This is not noise. The budget determines \emph{how many} sessions
get shrunk; the retention fraction, held at 12\%, determines \emph{how good each survivor is}
--- and the two budgets differ only in the first. To see what the second axis does we hold the
budget at its tightest (2~MB) and sweep the retention fraction directly (Table~\ref{tab:keepfrac}).

\begin{table}[t]
\centering
\caption{Return-consistency vs.\ retention fraction, 2\,MB budget, $S{=}16$, WorldMem. The
fraction sets how many of a session's $\approx$21 bank frames survive a shrink.}
\label{tab:keepfrac}
\begin{tabular}{lcccc}
\toprule
keep fraction & frames kept & anchor & lru & fifo \\
\midrule
0.04 & $\approx$1  & 0.6227 & 0.2127 & 0.1186 \\
0.08 & $\approx$2  & 0.6227 & 0.2127 & 0.1186 \\
0.12 & $\approx$3  & 0.6227 & 0.2127 & 0.1186 \\
0.25 & $\approx$5  & 0.6709 & 0.2085 & 0.1186 \\
0.50 & $\approx$10 & 0.6354 & 0.2085 & 0.1186 \\
\bottomrule
\end{tabular}
\end{table}

The result is stronger than a smooth trade-off, and it is the crux of the eviction story.
\textbf{Under relevance-keyed retention, keeping more frames barely helps} --- anchor reaches
0.62 while retaining a single frame, the anchor-relevant one, and adding nine more moves it only
to 0.64--0.67 (with a mild optimum near 25\%, past which drift-heavy middle frames re-enter and
slightly lower it). \textbf{Under recency-keyed retention, keeping more frames does not help at
all} --- lru is flat at 0.21 across the whole sweep. The two lines are parallel and never
converge.

This turns the relevance-versus-recency result from a difference of degree into a difference of
kind. Recency does not underperform because it retains too few frames; it underperforms because
it retains the \emph{wrong} frames --- the ones near where the rollout went, not the one it must
return to --- and no budget for keeping more of them closes the gap. A serving layer therefore
gains almost nothing from a larger retention fraction if it ranks by relevance, and nothing at
all from a larger one if it ranks by recency. The lever that matters is the ranking, not the
fraction.

\subsection{Spread across seeds}
\label{sec:eval-seeds}

\begin{table}[t]
\centering
\caption{Three seeds at 4\,MB, $S{=}16$.}
\label{tab:seeds}
\begin{tabular}{lccccc}
\toprule
policy & seed 0 & seed 100 & seed 200 & mean & sd \\
\midrule
pcs-aware & 0.9531 & 0.9595 & 0.9534 & 0.9553 & 0.0036 \\
anchor    & 0.6227 & 0.6219 & 0.6272 & 0.6239 & 0.0029 \\
lru       & 0.4110 & 0.4138 & 0.4407 & 0.4218 & 0.0164 \\
fifo      & 0.2981 & 0.3011 & 0.2986 & 0.2993 & 0.0016 \\
\bottomrule
\end{tabular}
\end{table}

The load-bearing quantity, anchor $-$ lru, is \textbf{+0.2021 $\pm$ 0.0136} --- an effect
roughly fifteen times the spread.

\paragraph{What this spread does and does not cover.} Seeds change what the model generates; they
do not change the runtime's control flow. \texttt{kept} and \texttt{drops} are \emph{identical}
across all three seeds (anchor 16/16 and 0; lru 11/16 and 10; fifo 5/16 and 11). So this is a
replication over generation randomness, not over scheduling. The runtime being deterministic is
itself a reportable property, but it means these intervals must not be read as covering
arrival-order or admission variability --- that is a separate replication, below.

\paragraph{Arrival-order replication.} We ran it. Varying the one thing the seed sweep holds
fixed --- the Poisson \emph{arrival} seed --- over eight arrival orders on the mock, under a
budget that binds ($S{=}24$, 4~MB; the arrival order alone moves peak concurrency to
$13.1 \pm 1.5$), the load-bearing quantity survives: \textbf{anchor $-$ lru $= +0.216 \pm 0.045$},
an effect $\approx$4.8$\times$ its arrival-order spread (the ranking never inverts). Two things
this exposes that the generation-seed sweep could not. First, arrival order is a \emph{larger}
source of variance than generation randomness ($\pm0.045$ vs $\pm0.014$), as it must be, since it
is what actually reorders eviction. Second, that variance is not shared equally:
\textbf{relevance-keyed retention is nearly immune to arrival order (anchor $0.5764 \pm 0.0003$)
while recency-keyed is markedly sensitive to it (lru $0.360 \pm 0.045$, fifo $0.336 \pm 0.062$)}
--- keeping the anchor is robust to \emph{when} sessions arrive, keeping the recent frames is not.

We then ran the same replication on the \textbf{real model}, since the scheduling counters are
model-independent but return-consistency is not. On WorldMem, over three arrival orders under a
binding budget ($S{=}8$, 4~MB, peak-live 8), the ranking not only holds but sharpens:
\textbf{anchor $0.697 \pm 0.025$, lru $0.362 \pm 0.062$, fifo $0.528 \pm 0.040$}, so
\textbf{anchor $-$ lru $= +0.335 \pm 0.084$}, again $\approx$4$\times$ its arrival-order spread and
never inverting across orders --- and again relevance-keyed retention is the more stable of the two
($\pm0.025$ vs $\pm0.062$). So the Section~\ref{sec:eval-evict} result --- relevance beats recency
under pressure --- is not an artifact of one arrival order on either the mock or the real model.

\subsection{Separating three things that were one knob}
\label{sec:eval-axes}

An earlier version of this evaluation reported a ``pcs-aware'' policy that scored 0.9531 at a
4~MB budget --- equal to the unconstrained baseline --- against 0.6227 for the anchor policy,
and the table appeared to be about eviction. It was not. That configuration varied three
independent decisions at once: how often to checkpoint, how to rank bank frames for eviction,
and which queued request to run next.

Separating the axes located the effect in the third. Under \texttt{resident-first} scheduling a
session runs its legs back to back, so it stays resident, and eviction --- which only targets
suspended sessions --- never reaches it \emph{during its own excursion}. The score was not good
eviction; it was \textbf{absence of eviction}. Two further observations confirm it: context
switches fall from 47 to 15 under that scheduler, and at a 2~MB budget the protection finally
fails (0.7756, three drops) because the pressure becomes impossible to dodge.

The checkpoint axis turned out to be inert on this workload: with legs of 4, 8 and 8 steps, the
cumulative step count is a multiple of the sparse policy's period at every request boundary, so
the sparse policy never actually skips a checkpoint. \texttt{every} and \texttt{sparse} produce
identical results, and any claim about checkpoint frequency would have to come from a workload
with irregular leg lengths.

The axes are now independent (\texttt{CheckpointPolicy} $\times$ \texttt{EvictionPolicy}
$\times$ \texttt{SchedulePolicy}), and the mapping from the original four names to the triples
they secretly were is recorded so earlier numbers remain interpretable.

\subsection{Trace-driven workload}
\label{sec:eval-trace}

The results above submit all requests at fixed times and force interleaving by construction.
That arranges the right \emph{situation} --- a session evicted mid-excursion --- rather than
causing it, and it also produces a pathological load: sixteen sessions opened at once, none ever
closing. The trace path replaces the arrangement with a cause. Sessions arrive by a Poisson
process, think between legs, and close; interleaving, queueing and memory pressure then follow
from arrival rate and think time. Two traces are used: a 16-session leave-and-return trace (80
events, 434~s span) and a planner trace exercising the fork path (152 events, 48 forks).

\begin{table}[t]
\centering
\caption{Leave-and-return trace, WorldMem, 2\,MB budget, all four policies.}
\label{tab:trace}
\setlength{\tabcolsep}{4.5pt}
\resizebox{\textwidth}{!}{%
\begin{tabular}{lrrrrrrrrr}
\toprule
policy & wall (s) & peak live & peak host (MB) & shrinks & drops & orphaned & returns scored &
  anchors lost & ret-consist \\
\midrule
anchor    & 632 & 6 & 2.30 & 14 & 0 & 0 & 16 & 0 & 0.8393 \\
fifo      & 633 & 6 & 2.30 & 23 & 8 & 0 & 16 & 0 & 0.8643 \\
lru       & 603 & 6 & 2.30 & 18 & 6 & 0 & 15 & 0 & 0.8900 \\
pcs-aware & 635 & 3 & 2.00 & 1  & 0 & 0 & 16 & 0 & 0.9531 \\
\bottomrule
\end{tabular}}
\end{table}

\paragraph{The pathology was in the workload, not the runtime.} Under real session lifetimes peak
concurrency is six live sessions, not sixteen, and no session's snapshots are ever lost --- every
policy scores its returns and \texttt{anchors lost} is zero throughout. The constructed load in
Section~\ref{sec:eval-evict} drove the anchor policy to 0.6227 at this budget; the same policy on
a realistic trace scores 0.8393, because sessions open and close rather than piling up. This is
the real-model confirmation of what the mock predicted, and it is the ground under
Section~\ref{sec:eval-neg}: predictive scheduling has nothing to optimise here because the budget
barely binds.

\paragraph{The eviction ranking still separates, but the gap is small and the sign is not what
Section~\ref{sec:eval-evict} would suggest.} Note that recency (lru, 0.8900) edges out relevance
(anchor, 0.8393) on this trace. This is not a contradiction; it is the flip side of the same
fact. When the budget barely binds, drops are rare (6 for lru, 0 for anchor) and the aggressive
shrinking that relevance-keyed retention performs to avoid drops is, here, unnecessary work that
trims frames no return needed. Relevance wins decisively when pressure is genuine
(Section~\ref{sec:eval-evict}, a 0.41 gap at hard pressure); it does not, and should not, win
when there is no pressure to manage. Reporting only the constructed load would have overstated
the policy's value; reporting only the trace would have understated it. Both belong.

\begin{table}[t]
\centering
\caption{Planner trace, fork path, WorldMem, anchor, 64\,MB.}
\label{tab:planner}
\begin{tabular}{lr}
\toprule
metric & value \\
\midrule
events & 152 \\
forks & 48 \\
peak live sessions & 23 \\
peak host (MB) & 11.5 \\
drops / orphaned & 0 / 0 \\
returns scored / lost & 4 / 0 \\
return-consistency & 0.8774 \\
\bottomrule
\end{tabular}
\end{table}

\paragraph{Copying fork holds on a real model.} The planner trace branches 48 times, each fork
cloning the parent PCS --- a real five-array bank, not a mock buffer --- and the four root
sessions all return with their anchors intact (\texttt{anchors lost = 0}, return-consistency
0.8774). This is the property the Section~\ref{sec:eval-scaling} correction turns on: because
WorldMem consumes its bank in place, a fork \emph{must} copy, and had the clone shared any array
the first branch to run would have corrupted the parent silently. Forty-eight forks with no lost
return is the evidence that it does not. Peak concurrency reaches 23 resident sessions here ---
higher than the leave-and-return trace, because a planner holds many branches open at once --- at
11.5~MB of host state, well within budget, so the fork path is exercised without eviction
confounding it.

\paragraph{Scoring under concurrent closes is correct.} Across both traces, every completed
return is scored and none is misattributed as a lost anchor, even though closes are releasing
snapshots throughout the run. This is the real-model check on the return-time scoring of
Section~\ref{sec:conformance}: the earlier post-hoc scorer would have reported near-total anchor
loss on exactly these workloads, because a trace, unlike the constructed load, actually closes
its sessions.

\paragraph{A real planner: MCTS.} The planner trace above is synthetic --- a fixed
\texttt{branch}-ary tree of fixed depth. A real search is not: its branching factor varies node
to node, its depths vary, and its pruning is visit-count driven. We therefore instrument a
genuine Monte-Carlo Tree Search over a navigation patrol --- go to a goal and return, the
smallest task that is both fork-and-backtrack (the search tree) and leave-and-return (the round
trip revisits the start viewpoint) --- and replay its tree through the runtime. What is real is
the control flow, which is what the serving layer sees: branching 1--3 node to node, depths
0--7, visit-count pruning, six committed patrols and 168 forks across the run; the world-model
legs use the same \texttt{Q}/\texttt{D}/\texttt{A} vocabulary, so it is a drop-in for the loads
above. This is the workload behind the counterfactual made concrete --- a planner time-sharing
one GPU across a fork-heavy search --- and it drives the eviction result to its extreme.

\begin{table}[tbp]
\centering
\caption{Real MCTS planner trace, WorldMem, by host budget. Footprint 30.3~MB;
\texttt{results/mcts\_trace\_worldmem\_*.json}.}
\label{tab:mcts}
\begin{tabular}{llcccc}
\toprule
host budget & policy & return-consistency & patrols returned & forks served (/168) & orphaned reqs \\
\midrule
64\,MB (unconstrained) & anchor & 0.8683 & 6/6 & 168 & 0 \\
64\,MB                 & lru    & 0.8683 & 6/6 & 168 & 0 \\
12\,MB & anchor & \textbf{0.8683} & \textbf{6/6} & \textbf{168} & \textbf{0} \\
12\,MB & lru    & \textbf{0.000}  & \textbf{0/6} & 144 & 48 \\
6\,MB  & anchor & \textbf{0.8535} & \textbf{6/6} & \textbf{168} & \textbf{0} \\
6\,MB  & lru    & \textbf{0.000}  & \textbf{0/6} & 103 & 130 \\
\bottomrule
\end{tabular}
\end{table}

Three readings, and together they are the eviction story in its starkest form.
\textbf{Without pressure the ranking does not exist:} at 64~MB, above the 30.3~MB footprint,
nothing is evicted, and relevance and recency are identical --- both return all six patrols at
0.8683. This is the control that the separation below is not an artifact of the policies doing
different amounts of work. \textbf{Under pressure, relevance is free and recency is fatal:}
relevance-keyed retention returns all six patrols at \emph{unchanged} quality (0.8683 at 12~MB
is exactly the unconstrained 0.8683, zero drops), while recency-keyed retention returns
\emph{none} (0/6). And \textbf{recency does not merely lose returns --- it corrupts the search:}
evicting by recency drops the shared ancestors the tree is still forking from, so 24 of 168
forks are orphaned at 12~MB and 65 at 6~MB (48 and 130 orphaned requests in all), and the deeper
the pressure the more of the planner's tree never gets built. Relevance orphans nothing at any
budget, because the anchor it protects is exactly the ancestor those forks depend on. On a
genuine planner workload the LLM-default eviction is therefore not suboptimal but unusable, and
the inversion this paper argues for is the difference between a search that completes and one
that collapses.

\subsection{Oversubscription: where the request-centric baseline is adequate --- and where it is not}
\label{sec:eval-oversub}

Section~\ref{sec:eval-scaling} argued the counterfactual --- that externalizing state is what lets
a session be suspended --- as arithmetic. Here we drive it as a workload, and it turns out to
sharpen the claim rather than simply confirm it. A persistent-world load (\texttt{gen\_oversubscribed}:
Poisson session arrivals that stay open across several away legs --- an RL-rollout-farm /
multi-client interactive-world shape) is replayed through the runtime twice, changing exactly one
thing: whether a suspended session's kernel is \textbf{checkpointed} (session-centric) or
\textbf{reclaimed} (request-centric --- keep only the cheap last observation, drop the bank and the
RNG: the incomplete $H'$ of Section~\ref{sec:gap}). Arrival rate sets the offered concurrency; we
report the \emph{measured} peak-live count, since that is the concurrency the single
time-multiplexed GPU actually saw.

\begin{figure}[tbp]
\centering
\includegraphics[width=0.82\linewidth]{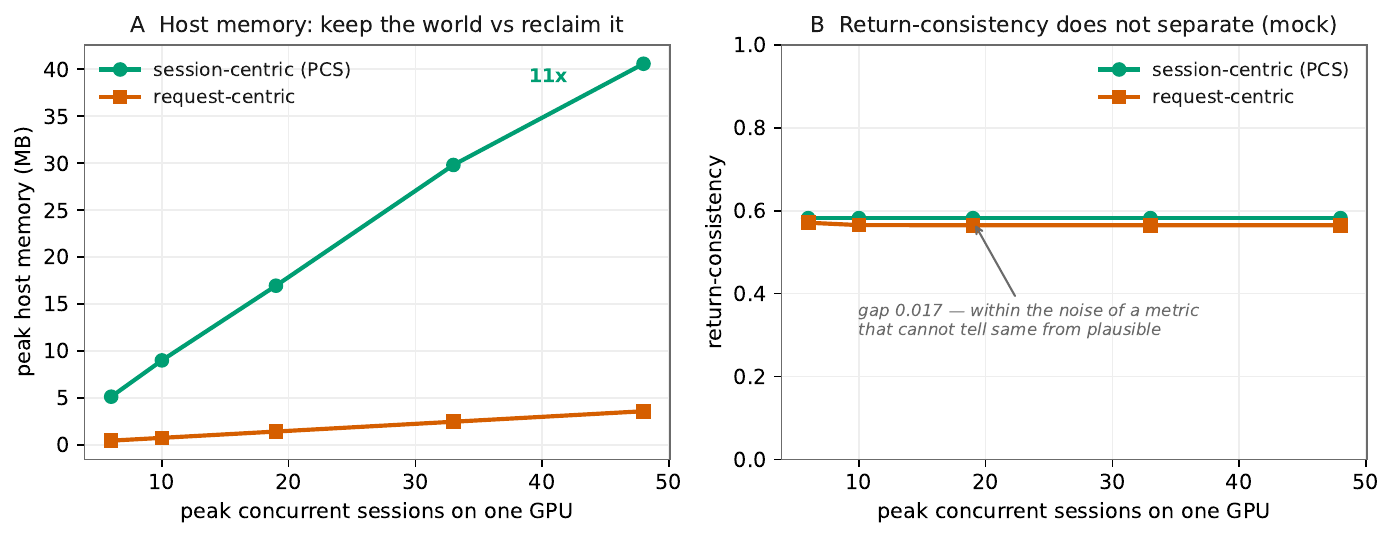}
\caption{\textbf{Oversubscription, mock sweep.} As peak concurrent sessions on one
time-multiplexed GPU rise, the session-centric runtime's host memory grows steeply (it keeps every
session's world) while the request-centric server's grows an order of magnitude slower (it keeps
only the last frame) --- $\approx$11$\times$ at 48 sessions. Return-consistency does not separate
(right): request-centric is adequate on this recency-dominated model. The advantage it forfeits is
not semantic but durability (Section~\ref{sec:eval-c3}).}
\label{fig:oversub}
\end{figure}

\paragraph{The systems behaviour is exactly the counterfactual.} As the load oversubscribes one
GPU --- peak concurrency rising to 48 resident sessions --- the session-centric runtime's host
memory grows steeply with concurrency (it keeps every session's world) while the request-centric
server's grows an order of magnitude more slowly (it keeps only the last frame, discarding the
accumulated world at each boundary): 40.6~MB versus 3.6~MB at 48 sessions, $\approx$\textbf{11$\times$}
(Figure~\ref{fig:oversub}). The real model at a smaller point (peak-live 8) lands at the same ratio,
7.2~MB versus 0.59~MB ($\approx$12$\times$) --- mock and real agree. This is the
Section~\ref{sec:eval-scaling} projection turned into a measured curve, and it is unambiguous: only
externalized state can be suspended.

\paragraph{The semantic metric, however, does not separate --- and we report that plainly, because
it is the interesting part.} Return-consistency barely moves between the two: on the mock the
request-centric server scores 0.565 to the session-centric 0.582 (a 0.017 deficit), and on the
\textbf{real WorldMem it scores 0.60 to the session-centric 0.58 --- marginally \emph{higher}, at
12$\times$ less host memory, with no anchor lost by either}. The reason is the one
Section~\ref{sec:eval-trace} already exposed: this model is recency-dominated. Its leave-and-return
re-grounds on pose and on recent frames, not on the accumulated anchor-side bank, so carrying the
whole bank across a switch buys nothing here --- and the single most-recent frame the
request-centric server keeps is, if anything, a cleaner conditioning context than a bank thick with
stale away-frames. Pushed to the extreme --- one frame versus the whole bank --- this is the same
result as Section~\ref{sec:eval-trace}'s low-pressure row, where recency edged relevance. We could
have manufactured a return that only the anchor bank can serve, but on a recency-dominated model
that would be constructing the effect rather than measuring it.

\paragraph{So the session-centric advantage under oversubscription is not semantic
return-consistency --- it is durability.} A request-centric resume is \emph{plausible}, not
\emph{identical}: having dropped the RNG it cannot reproduce the world it left, only a world
consistent with the last frame it kept. The return-consistency metric cannot see this, because both
return a world that \emph{looks} right; \textbf{bit-exactness can, and it is exactly the property
Section~\ref{sec:eval-c3} measured request-centric to forfeit} --- the cross-process restore is
byte-identical \emph{because} the RNG and the kernel were kept, and a server that reclaims them at
the boundary structurally cannot match it. For the usage this paper is about --- a planner or
trainer that forks candidate rollouts and backtracks --- that gap is decisive rather than cosmetic:
candidates continued from a drifted, non-deterministic restore are not comparable to one another,
and a backtrack does not return to the state it left. The requirement there is bit-exact restore,
and only the session-centric runtime meets it.

This bounds the claim precisely, which is worth more than an inflated version of it. Request-centric
serving loses the world \emph{when the model's memory is load-bearing for the return} --- which
WorldMem's, on this protocol, is not. What request-centric cannot do on \textbf{any} model is
return the world \emph{the same} rather than merely plausible, and that byte-level identity, not
semantic similarity, is what a fork-and-backtrack workload actually consumes. The oversubscription
study is therefore not evidence that request-centric drifts; it is evidence that the primitive it
lacks is durability, and that a semantic acceptance test alone would have hidden the gap.

\subsection{What did not work}
\label{sec:eval-neg}

Two designs were built, measured, and rejected. Both are reported because the reason they fail
bounds the claim the runtime can make.

\paragraph{Predictive eviction.} The hypothesis was that shrinking while slack remains avoids the
regime where the only remaining move is deletion. It does the opposite, monotonically: lowering
the trigger from 100\% of budget to 50\% takes drops from 20 to 26 and worlds kept from 6/16 to
3/16. The runtime escalates shrink $\to$ nothing-left-to-shrink $\to$ drop, so evicting to a
lower target pushes more sessions to the bottom of their shrink range and the next round has no
option but deletion. The reactive path's laziness is protective. The harm is eviction
\emph{volume}, not eviction \emph{timing}, and no scheduling of eviction addresses it.

\paragraph{Predictive admission.} If the problem is upstream, gate it upstream: project host
demand over declared arrivals and delay or refuse sessions that cannot be kept whole. On the
trace workload the reactive baseline already completes 16/16 worlds with zero drops, because real
session lifetimes keep peak concurrency near six and the budget never binds. Every admission
setting therefore refuses worlds that would have completed: 5/16 intact at 95\% safety, 2/16 at
the most aggressive.

Both optimise a regime this workload never enters --- but Section~\ref{sec:eval-oversub} built one
that does, and we replayed both designs on it rather than leave the question open. The
persistent-world load there genuinely oversubscribes the single GPU (peak concurrency 29 of 48
sessions at a 1~MB budget), and neither predictive design pays. \textbf{Predictive eviction is
exactly inert}: at every budget and every trigger it produces byte-for-byte the same result as
reactive (39 drops at 1.5~MB, 67 at 1~MB), because once every bank is shrunk to its 12\% retention
floor the only move left is to drop, and firing earlier cannot shrink below the floor --- the floor
caps what eviction can save, whoever schedules it. \textbf{Predictive admission is harmful at any
usable operating point and marginal only past catastrophe}: at moderate pressure it refuses worlds
that would have completed (6/48 admitted where reactive completes 9), and only where reactive has
already collapsed to 1 of 48 does refusing 41 sessions to protect 4 edge ahead ($+0.036$) --- a
salvage, not an operating point. So the falsification is not an artifact of an unstressed load:
under genuine oversubscription the reactive path is still the right one, for a sharper reason than
before --- the retention floor, not the arrival rate, is what bounds the gain. Both implementations
are retained (\texttt{session\_runtime/predictive.py}, \texttt{predictive\_v2.py}; the
oversubscription replay is \texttt{run\_predictive\_oversub.py},
\texttt{results/predictive\_oversub\_mock.json}) with their ablation tables.

\emph{A note on the third component.} Prefetching a session's state before its request arrives
achieved a 100\% hit rate and no measurable benefit, exactly as predicted: restore costs
0.012~ms against a 1.85~s step, so there is no stall to hide. It was implemented because an
ablation in which every idea works is an ablation that was not run.

\subsection{Threats to validity}
\label{sec:eval-threats}

\begin{itemize}[leftmargin=1.4em]
\item \textbf{One GPU, time-multiplexed.} Constant device memory is a property of this design,
  not of world-model serving in general. A concurrent multi-GPU runtime is a different system and
  its numbers would not be comparable to these.
\item \textbf{Per-session host cost is not a constant.} 0.964~MB/session here versus 0.259~MB in
  the earlier single-snapshot measurement; it scales with horizon $\times$ snapshots retained.
\item \textbf{Exactness is shown on a Markovian model.} The same discovery procedure extends to
  the non-Markovian one, and restore is bit-exact there too, but the claim that the observation
  \emph{is} the complete visible state is specific to the Markovian case.
\item \textbf{Seeds cover generation only} (Section~\ref{sec:eval-seeds}).
\item \textbf{No positional-injection result, and the substrate explains why.} An experiment
  intended to test whether ``frozen'' and ``unreachable'' are two ends of one injection-position
  axis returned one distinct result across six placements: the position parameter had no effect,
  so the hypothesis was never exercised and the result file is marked invalid rather than
  reported. What the failure did establish is a constraint --- this model requires
  \texttt{num\_frames} $\in \{$chunk, chunk$+1\}$, so its frame budget is pinned to the action
  chunk and there is no position to assign. That is consistent with the mechanism concurrent work
  relies on: addressable-memory schemes reposition entries within a temporally-encoded KV cache,
  which a chunk-conditioned diffusion model does not expose. The question is well posed only on a
  model with an addressable memory channel. The third model in this paper, Matrix-Game 2.0, is
  exactly such a substrate: its PCS \emph{is} a temporally-encoded KV cache, and each of its 30
  blocks carries explicit \texttt{global\_end\_index} / \texttt{local\_end\_index} position
  indices (captured as part of the fingerprint, Table~\ref{tab:pcs}). We ran the experiment on it,
  and report two findings. \emph{First}, arbitrary-position injection is blocked by the model's own
  cache manager: its windowed KV cache assumes \emph{contiguous} positions (each write index is the
  previous end), so injecting an entry at a non-contiguous offset overruns the window buffer. The
  addressability the substrate exposes is read-only, not write-repositionable without rewriting the
  cache manager. \emph{Second}, and more informative, we measured whether content becomes
  \emph{unreachable} past the local-attention window (six frames) with a causal probe --- two
  rollouts differing only in the initial frame, driven by identical actions, with the divergence
  between them tracking how long the initial frame still determines the output. It does not become
  unreachable: with the camera held, divergence \emph{grows} to $1.48\times$ and plateaus through 27
  frames ($4.5\times$ the window); and a leave-and-return excursion that pans the start out of view
  keeps the two returns divergent (return divergence $\ge$ anchor divergence) out to 18-frame
  excursions ($3\times$ the window). The finite window bounds direct attention but not effective
  memory, which the autoregressive chain carries forward. So the frozen-vs-unreachable
  position axis the earlier experiment sought does not cleanly manifest on this model --- not because
  the substrate cannot pose the question, but because propagation dominates the window. This is a
  \emph{measured} negative (\texttt{results/mg2\_\{reachability,excursion\}.json}), the substrate
  Cosmos3 could not provide, exercised.
\end{itemize}

\section{Implementation}
\label{sec:impl}

The runtime is a single-process Python service, \texttt{session\_runtime}, of roughly a thousand
lines: a session table, admission control, a request queue, a pluggable policy layer, and a PCS
store. Model access is confined to one adapter interface of five methods
(\texttt{initial\_state}, \texttt{step}, \texttt{capture}, \texttt{restore}, \texttt{embed}),
which is what allowed the entire service --- policy logic, queueing behaviour, the shape of the
scaling curves --- to be developed against a mock world model on a laptop, with only the final
measurements bound to a real model on a GPU. The clock is injected for the same reason: a virtual
clock for development, a real clock for anything reported.

WMMS, the design this work grew out of, is one realization of the same contract at a different
level of ambition; it adds identity, ownership and consolidation semantics that this paper
deliberately does not claim. The contributions here --- PCS, discovery by measurement, the
conformance test --- are independent of both implementations and apply to any engine exposing the
three fingerprintable operations of Section~\ref{sec:discover}.

Everything is reproducible from \texttt{experiments/README.md}, which maps each script to the
JSON it produces and the claim that JSON supports.

\section{Related Work}
\label{sec:related}

\paragraph{Benchmarks that established the failure --- and what they concluded from it.} Within
eight weeks of each other, four benchmarks independently measured long-horizon memory in video
world models at a scale no single group had attempted before. MBench decomposes memory into
twelve sub-dimensions over fourteen models and finds that under departure-and-return camera
trajectories ``the scene does not return to the same 3D configuration even when the camera motion
semantically suggests a departure-return trajectory.'' WRBench treats camera motion as an
intervention on observability, evaluates 23 models across four control paradigms on 9{,}600 videos
calibrated by 2{,}547 human verdicts, and reports a failure it calls stubborn: systems ``maintain
the observed world as a tracking shot.'' MemoBench~\cite{memobench} isolates the same property through a
disappear-and-reappear protocol, and CoW-Bench~\cite{cowbench} folds it into a trinity of modal, spatial and
temporal consistency. A concurrent roadmap article~\cite{roadmap} sharpens why this matters by \emph{defining} a
simulator --- as distinct from a renderer --- as a model that must preserve ``geometry, physical
constraints, object persistence, and action-conditioned temporal regularities sufficiently well
to support reliable forward prediction.''

\paragraph{We take that finding as established and do not re-prove it.} These benchmarks
demonstrate the phenomenon at a scale we cannot match, and this paper would be weaker, not
stronger, for adding a fifth measurement of it. What we dispute is the \emph{attribution}. Every
one of them locates the deficiency in the model: WRBench concludes that progress requires ``not
more pixels, but a what-memory that records hidden change and a training objective that supervises
endpoint persistence,'' and reads its scaling evidence --- Wan's re-observed state falling from
0.66 to 0.62 as parameters grow from 1.3B to 14B --- as showing that capacity cannot supply it.
For the model class we examine, that attribution is wrong in a way that is directly demonstrable:
a \texttt{uint8} observation plus the generator's RNG state, snapshotted by the runtime and
restored after an intervening excursion, reproduces the never-left continuation
\emph{byte-identically} (Section~\ref{sec:eval-c3}). The capability was never absent; the runtime
discarded the state it needed. The same roadmap that defines the simulator requirement also notes,
in its survey of evaluation, that ``existing benchmarks often overlook inference efficiency'' ---
which is the layer this paper is about.

\paragraph{A note on the word ``persistence''.} WRBench uses it for the requirement that an
\emph{unobserved} world keep evolving --- the moon continues its orbit, an event set in motion runs
to its conclusion --- so its failure mode is a world frozen at the moment observation stopped. We
use it in the computational sense: state that should survive a request boundary in fact survives
it, so our failure mode is a world that cannot be recovered at all. These are two halves of one
deficiency, since a system with no state object neither advances what should advance nor preserves
what should be preserved, but they are not interchangeable and a reader who conflates them will
misread both.

\paragraph{Addressable memory within an episode.} Closest to this work in spirit, and orthogonal
to it in layer, is WorldTrace~\cite{worldtrace}, which asks why long-horizon persistence degrades in autoregressive
video world models and answers: not because past content is absent from the KV cache, but because
it becomes \emph{unaddressable}. Once a rollout exceeds the training horizon, the temporal RoPE
offset to a distant cached frame falls outside the range seen during training, and the model can
no longer attend to an entry that is still physically present. A second, coupled failure is that
naive compression of RoPE-rotated keys averages together incompatible positional phases, so the
summary that survives is uninformative even when it is reachable. WorldTrace resolves both without
retraining, by assigning each summary slot a virtual position derived from its \emph{slot rank}
rather than its original timestamp --- keeping every slot in-distribution at any horizon and
distinguishable from its neighbours --- and by performing compression in the canonical, unrotated
key domain. Its two writers target different queries (averaged fields for temporal coherence,
frozen verbatim landmark keys for episodic recall), and it is evaluated on LoopMem, a controlled
scene-revisit benchmark, where it improves temporal consistency by 15.5\% and episodic recall by
19.5\% on Matrix-Game-2.

The relationship to this paper is worth stating precisely, because the abstracts sound similar.
\textbf{WorldTrace makes a model's own memory usable up to its capacity; we are concerned with
what happens when the session ends.} It is a per-model mechanism operating \emph{inside} one
rollout's attention window, and it presupposes a model whose memory lives in a temporally-addressed
KV cache. It does not address survival across a request or process boundary, forking, sharing
between clients, or reuse by a different model --- and a model that exposes only a state-port, with
no addressable memory channel, cannot host it at all. We observed exactly that boundary
empirically: the diffusion model we evaluate constrains its frame budget to the action chunk,
exposing no positional knob for injected memory (Section~\ref{sec:eval-threats}), which is why the
addressability question is not one our substrate can even pose. The two results compose rather than
compete: WorldTrace determines whether retrieved memory is \emph{readable}, PCS determines whether
it still \emph{exists}.

One of its findings does constrain our design, and we adopt it as a scope boundary. Because
compression in the rotated domain silently cancels phase, any eviction that \emph{merges} cached
entries must operate in canonical space or corrupt what it summarises. Our eviction selects frames
rather than merging them --- it retains a subset and discards the rest --- so the phase-cancellation
failure mode does not arise here. A consolidating eviction policy would inherit the constraint, and
we note it rather than leave it implicit.

\begin{figure}[tbp]
\centering
\includegraphics[width=0.92\linewidth]{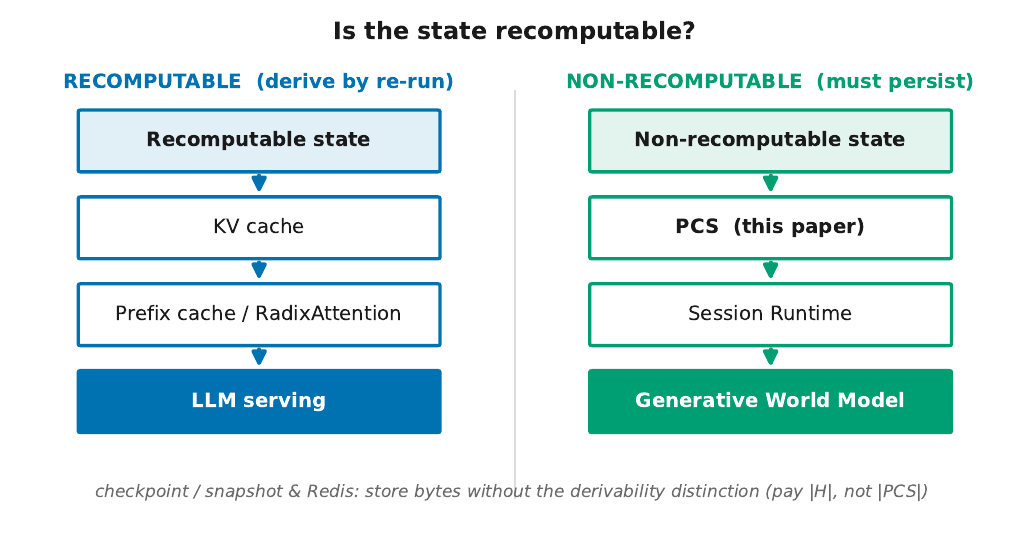}
\caption{\textbf{A taxonomy by derivability.} \emph{Is the state recomputable?} Recomputable state
(KV cache, prefix cache / RadixAttention, LLM serving) may be discarded and rebuilt.
Non-recomputable state (PCS, this paper's session runtime, generative world models) must be
persisted. Checkpoint/snapshot and Redis store bytes without the derivability distinction --- they
pay $|H|$, not $|\mathrm{PCS}|$.}
\label{fig:taxonomy}
\end{figure}

\paragraph{A taxonomy by derivability.} The cleanest way to place the abstraction is along a single
axis: \emph{is the state recomputable?} (Figure~\ref{fig:taxonomy}). On one side, recomputable
state is derivable by re-running compute --- KV caches, prefix caching, RadixAttention~\cite{sglang},
tree-sharing, and the KV-compression, streaming, and layer-wise management schemes built on
them~\cite{cachegen,cacheblend,layerkv} --- all correctly treated as optimizations that may be
discarded and rebuilt. On the
other, non-recomputable state carries a kernel that no re-run reconstructs, and managing it means
\emph{persisting} it. Prior state-management primitives sit at fixed points on this axis and miss
the distinction PCS is built on. A KV cache stores only recomputable state and assumes recompute is
always available. A checkpoint or snapshot facility --- OS, database, \texttt{CRIU}-style process
image~\cite{criu} --- stores bytes without asking what is recomputable: correct, but it pays $|H|$ rather than
$|\mathrm{PCS}|$. A durable key-value store offers bytes and eviction but no notion of derivability
and no semantic acceptance criterion, and its default recency eviction is not merely suboptimal for
world memory but \emph{wrong} (Section~\ref{sec:eval-evict}). PCS is the point these miss: the
minimal set on the non-recomputable side, discovered by measurement rather than asserted, and
validated semantically rather than by byte equality.

\paragraph{Generative serving mechanism (foundation, not competitor).} Paged-KV~\cite{vllm} and
autoregressive-diffusion serving engines, iteration-level and disaggregated schedulers~\cite{orca,sarathi,distserve},
production inference stacks~\cite{tensorrtllm}, typed session-memory managers, and copy-on-write KV fork
supply the mechanism this semantic layer stands on --- how to store, page, evict and fork bytes
cheaply. We consume these directly; the contribution is the layer above them, and the division is
deliberate: bytes are somebody else's problem, and should be.

\paragraph{Prefix caching and RadixAttention~\cite{sglang} --- the clean differentiator.} These reuse a prefix's
computation under an assumption that is load-bearing and usually invisible: that the cached state
is \emph{deterministically recomputable} by re-running the same tokens. World models violate it.
They carry a non-derivable kernel --- the sampling RNG, and in non-Markovian models the memory bank
or recurrent activations --- that no amount of re-running reconstructs, because the noise that
produced a particular world is not a function of the inputs. A recompute-caching runtime is
therefore not merely inefficient for this model class; it is \emph{incorrect}, and silently so,
since the reconstructed state is a plausible world rather than the one that was left. This is the
one-paragraph reason the abstraction is new rather than a rebranding of prefix caching.

\paragraph{Externalized memory where language suffices.} A separate line externalizes model
\emph{knowledge} into a queryable memory model addressed in natural language, keeping the executive
model frozen and opaque. Its advantages --- cross-model transfer, no representational coupling,
plug-in replacement --- follow from the interface: text is a model-agnostic, losslessly
transferable format for facts. World \emph{state} has no such interface. Exact geometry, appearance
and dynamics cannot be written down in a form that lets any model re-ground on them identically; a
scene regenerated from a description is one that \emph{matches the description}, with everything the
words did not pin re-imagined. That fidelity-versus-transferability trade is the boundary between
the two problems: language-serializable knowledge can take the decoupled route, and world state
cannot, which is why it needs a system of its own rather than a shared one.

\paragraph{Memory architectures our test can score.} Explicit-memory world models --- pose-keyed
banks, retrieval-augmented caches, the addressable schemes above --- are \emph{methods}. Our
conformance criterion (Section~\ref{sec:conformance}) accepts or rejects a restore they produce; it
does not compete with them. One of them serves here as a positive control precisely because it has
memory to find.

\section{Discussion, limitations, and conclusion}
\label{sec:conclusion}

\paragraph{Necessity, not convenience.} A reader may grant that PCS works and still ask whether
it is \emph{necessary} --- whether a runtime could get by with a generic checkpoint feature and
good defaults. It cannot, for a reason specific to this state, and the argument is worth making
because it is what separates an abstraction from a utility. The two obvious alternatives to PCS
are not slower versions of it; they are incorrect or unbounded. \emph{Recompute-caching} --- the
assumption LLM serving is built on --- does not merely underperform here: because the sampling
RNG and the memory kernel are non-derivable, re-running the inputs reconstructs a \emph{different,
plausible} world, so the runtime returns the wrong answer and reports success. No tuning of a
recompute cache fixes this, because the fault is in the assumption, not the configuration.
\emph{Snapshotting the entire runtime state} is correct but pays $|H|$ per session and grows with
horizon, so it is not a serving primitive but a schedule for running out of memory. PCS is the
only point that is both correct --- it keeps exactly the non-derivable kernel --- and bounded ---
it keeps nothing else. The derivability classification is therefore load-bearing, not cosmetic: a
runtime that does not separate derivable from non-derivable state cannot decide what to keep
without being either wrong or unbounded. Necessity here is not that PCS is the most convenient
abstraction available; it is that the correct-and-bounded region has exactly one occupant.

\paragraph{Limitations.}
\begin{enumerate}[leftmargin=1.6em]
\item \textbf{The fingerprint is a procedure, not a theorem.} Existence is constructive;
  uniqueness, stability under retraining, and probe-independence are open and stated as scope
  (Section~\ref{sec:discover}).
\item \textbf{Three model families, not a universality proof.} Cosmos3, WorldMem, and
  Matrix-Game 2.0 sample three distinct memory architectures --- Markovian observation, growing
  non-Markovian bank, and windowed addressable KV context --- and the fingerprint returns a
  structurally different PCS on each (constant / growing / windowed), restored byte-identically
  on all three including across a process boundary. Three points are a genuine sampling of the
  design space, but not a proof of universality; architectures we did not test --- recurrent /
  RSSM latent-state models~\cite{planet,dreamerv3}, other action-conditioned generative world
  models~\cite{genie,gaia1}, and external retrieval memories --- remain open.
\item \textbf{One GPU, time-multiplexed.} Constant device memory is a property of this design, not
  of world-model serving in general (Section~\ref{sec:runtime}).
\item \textbf{Per-session cost is not a constant.} It scales with horizon $\times$ snapshots
  retained; the figure quoted for one configuration does not transfer to another
  (Section~\ref{sec:eval-threats}).
\item \textbf{The retention fraction, not the budget, sets quality.} For the relevance-keyed policy
  the budget determines how many sessions get pruned while the retention fraction determines how
  good each survivor is, which makes the cost curve a step function in that policy's case
  (Section~\ref{sec:eval-evict}).
\item \textbf{Replication covers generation, not scheduling.} Seeds change what the model generates
  but not the runtime's control flow, so the reported spread does not cover arrival-order
  variability (Section~\ref{sec:eval-seeds}).
\item \textbf{No predictive scheduling result.} Two designs were falsified on the
  reactive-friendly workload \emph{and} on a genuinely oversubscribed one
  (Section~\ref{sec:eval-neg}): predictive eviction is bounded by the retention floor, predictive
  admission helps only past collapse. The runtime claims no scheduler.
\item \textbf{Metric-dependence.} The tolerance $\epsilon$ and the persistence horizon are defined
  relative to the return-consistency metric.
\item \textbf{Model-bounded persistence.} The runtime preserves exactly the persistence the model
  has. No runtime can manufacture memory the model never stored.
\end{enumerate}

None of these gate the core claims; they bound their scope.

\paragraph{Conclusion.} Four benchmarks established in 2026 that video world models lose the world
across an excursion, and each concluded that the remedy is a better model. For an important class of
models that conclusion is wrong, and the demonstration is direct: a \texttt{uint8} observation plus
the generator's state, snapshotted by the runtime and restored after genuine intervening work,
reproduces the never-left continuation byte-identically. The capability was never missing. The
runtime threw the state away.

What follows is an abstraction rather than a patch. The unit of serving for generative world models
should move from the \textbf{request} to the \textbf{session}, and the object a session owns is
\textbf{Persistent Computational State} --- minimal, non-recomputable, and, crucially,
\emph{measurable} rather than asserted. The same discovery procedure returns two structurally
different answers on three models --- flat, growing, and windowed --- and restores all three exactly.
Holding such state is not the hard part;
at 0.012~ms against a 1.85~s generation step it is free. The hard part was semantic: \emph{what} to
save, \emph{what to drop first} when memory binds --- where the answer inverts LLM practice --- and
\emph{how to know a restore worked}, which byte equality cannot tell you.


\end{document}